\pdfoutput=1

\documentclass[11pt]{article}

\usepackage[final]{acl}

\usepackage{times}
\usepackage{latexsym}
\usepackage{siunitx}
\usepackage[T1]{fontenc}

\usepackage[utf8]{inputenc}

\usepackage{microtype}

\usepackage{inconsolata}

\usepackage{graphicx}

\usepackage{verbatim}
\usepackage{amsmath}
\usepackage{colortbl}
\usepackage{todonotes}
\usepackage{xspace}
\usepackage{booktabs}
\usepackage{multirow}
\usepackage{diagbox}

\newcommand{\mirage}{\textsc{MiRAGE}\xspace}
\newcommand{\frameworkname}{\mirage}

\newcommand{\claimscore}{\textsc{InfoF1}\xspace}
\newcommand{\claimprec}{\textsc{InfoP}\xspace}
\newcommand{\claimrec}{\textsc{InfoR}\xspace}
\newcommand{\citationscore}{\textsc{CiteF1}\xspace}
\newcommand{\citeprec}{\textsc{CiteP}\xspace}
\newcommand{\citerec}{\textsc{CiteR}\xspace}

\newcommand{\autoargue}{\textsc{Auto-ARGUE}\xspace}
\newcommand{\argue}{\textsc{ARGUE}\xspace}
\newcommand{\acle}{\textsc{ALCE}\xspace}
\newcommand{\ragas}{\textsc{RAGAs}\xspace}
\newcommand{\wikivideo}{\textsc{WikiVideo}\xspace}
\newcommand{\clue}{\textsc{CLUE}\xspace}
\newcommand{\unli}{\textsc{UNLI}\xspace}
\newcommand{\clotho}{\textsc{Clotho}\xspace}

\title{
Seeing Through the \frameworkname: \\Evaluating Multimodal Retrieval Augmented Generation 
}

\author{
Alexander Martin\textsuperscript{\rm 1}  
\quad William Walden\textsuperscript{\rm 1,2*}  
\quad Reno Kriz\textsuperscript{\rm 1,2*} 
\quad Dengjia Zhang\textsuperscript{\rm 1} \\
\quad \textbf{Kate Sanders}\textsuperscript{\rm 1}
\quad \textbf{Eugene Yang}\textsuperscript{\rm 1,2}
\quad \textbf{Chihsheng Jin} 
\quad \textbf{Benjamin Van Durme}\textsuperscript{\rm 1,2}
\\
  \textsuperscript{1}Johns Hopkins University\quad \textsuperscript{2}Human Language Technology Center of Excellence\quad \\
  \texttt{\small{\{amart233, vandurme\}@jhu.edu}}
}

\begin{document}
\maketitle
\begin{abstract}

We introduce \mirage, an evaluation framework for retrieval-augmented generation (RAG) from multimodal sources. As audiovisual media becomes a more prevalent source of information online, RAG systems must integrate such media into generation. Yet, existing evaluation methods for RAG are largely text-centric and do not readily transfer to multimodal settings. \mirage is a claim-centric approach to multimodal RAG evaluation, consisting of \claimscore, which assesses factuality and information coverage, and \citationscore, which assesses citation support and completeness. We show that, when applied by humans, \mirage strongly aligns with extrinsic judgments of output quality. We additionally introduce an automatic implementation of \mirage and compare it to multimodal variants of three prominent text-centric RAG metrics---\acle, \argue, and \ragas---finding that \mirage outperforms all three on text while being the only one to generalize to multimodal sources. We release open-source implementations and outline evaluation methods for multimodal RAG.\footnote{\url{https://github.com/alexmartin1722/mirage}}

\end{abstract}

\section{Introduction}

\begin{figure}
    \centering
    \includegraphics[width=\linewidth]{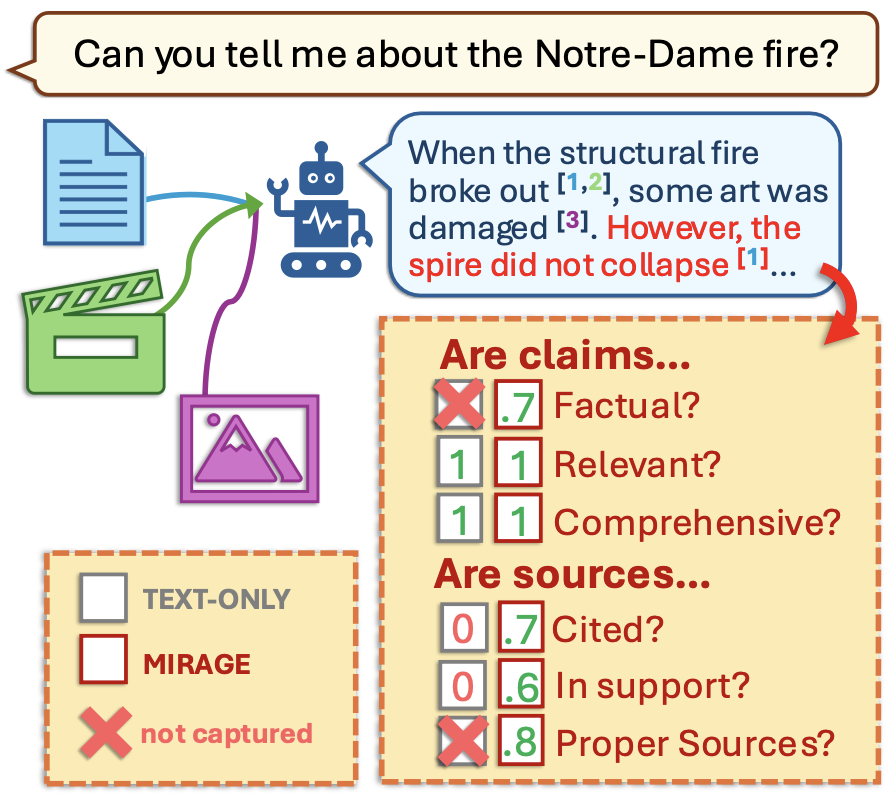}
    \caption{\frameworkname assesses predictions at the subclaim level, evaluating information factuality and coverage, as well as citation support and attribution, enabling RAG evaluation grounded in text, image, audio, and video.}
    \label{fig:teaser}
    \vspace{-1em}
\end{figure}

Online information is increasingly consumed through audiovisual media---from firsthand video footage of natural disasters to professional news coverage of major political events. It is thus essential that systems for retrieval-augmented generation~(RAG) be able to integrate information from \emph{any} modality and provide accurate citations to multimodal sources. However, work on long-form text generation from multimodal sources \cite{krishna2017dense, liu2023visualinstructiontuning, lin2024videoxumcrossmodalvisualtextural, martin-etal-2026-wikivideo} relies on text-centric metrics for evaluation \cite[e.g.,][]{lin-2004-rouge, zhang-etal-2019-bertscore}, which do not verify information against non-text sources.

Subclaim-level factuality metrics like FActScore \cite{min-etal-2023-factscore} verify outputs by decomposing them into subclaims and verifying each subclaim against a knowledge source. However, FActScore and its variants \cite{song-etal-2024-veriscore} are designed for single-source, text-to-text verification, and do not handle citations, multiple sources, or multimodal evidence. RAG evaluation requires more: systems must not only produce factual text, but attribute information to the correct sources, which may be images, audio, or video, in addition to text.

Existing text-only RAG metrics---e.g., \acle \cite{gao-etal-2023-enabling}, \argue \cite{mayfield2024nuggets}, and \ragas \cite{es-etal-2024-ragas}---address citation quality and information coverage, but are designed for text passages and operate at the sentence level, precluding fine-grained (i.e., subclaim-level) factuality judgments. Extending these metrics to multimodal sources is not straightforward: multimodal inference introduces a much heavier computational burden (e.g., fitting multiple videos in context), novel challenges relating to evidential support (e.g., how to comprehensively extract facts from videos), and (sub)claim grounding that requires cross-modal reasoning over visual and auditory signals. We show that naively adapting text-only RAG metrics to video yields poor correlation with human judgments (\S\ref{section:human-alignment}), confirming the need for a framework designed specifically for multimodal RAG.

We introduce \mirage, a framework for \textbf{M}ult\textbf{i}modal \textbf{R}etrieval \textbf{A}ugmented \textbf{G}eneration \textbf{E}valuation that extends subclaim-level factuality assessment to the multimodal, multi-source, citation-attributed setting. Core to \mirage is the notion that all information---whether represented in text, image, audio, or video---can be decomposed into subclaims, grounding all modalities in a common unit of evaluation. Our framework consists of \claimscore, which measures the factuality (precision) and coverage (recall) of information in generated text, and \citationscore, which measures whether cited sources support their associated subclaims (precision) and whether recalled information is properly attributed (recall). Both metrics can be evaluated in reference-based or reference-free settings, enabling \mirage to be used across many different tasks (e.g., summarization, visual question answering).

We evaluate on three RAG datasets spanning text and audiovisual sources: RAGTIME \cite{lawrie2026overviewtrec2025ragtime}---a multilingual text report-generation task---and \wikivideo \cite{martin-etal-2026-wikivideo} and \wikivideo-25 \cite{martin2026findingsmagmar2026shared}---two audiovisual article-generation tasks. Comparing against three types of human judgments, we find that \mirage has the highest agreement with human judgments across all three datasets and modalities, outperforming the text-native metrics on text and generalizing to other modality sources where they do not. On audiovisual sources, subclaim-level precision (\claimprec, \citeprec) is the only automatic metric that consistently correlates with grounding judgments, while other metrics have low or negative correlation.

In summary: (1) We propose \mirage, extending subclaim-level factuality assessment to multimodal, multi-source, citation-attributed RAG; (2) we present multimodal versions of three prominent text-only RAG metrics, revealing their limitations; and (3) we evaluate across three datasets and three types of human judgments, showing that \mirage outperforms the text-native metrics on text \emph{and} is the only metric to generalize to other modalities---extending reliable evaluation across modalities at no cost to text-domain performance.

\section{Related Work}

\paragraph{Retrieval-Augmented Generation} has been widely studied in text-only settings \citep[TextRAG;][]{lewis2020retrieval,barham2023megawikamillionsreportssources, gupta2024overviewtrec2024biomedical, han-etal-2024-rag, lawrie2024overviewtrec2023neuclir}. Recent work has extended RAG to other modalities, including documents with visual elements \cite{cho2024m3docragmultimodalretrievalneed}, speech \cite{min2025speechretrievalaugmentedgenerationautomatic, chen-etal-2025-wavrag}, and video \citep[VideoRAG;][]{jeong2025videoragretrievalaugmentedgenerationvideo, martin-etal-2026-wikivideo, ren2025videoragretrievalaugmentedgenerationextreme}. We focus on the VideoRAG task as proposed in \citet{martin-etal-2026-wikivideo}, as it best mirrors the common TextRAG setting: long-form generation from multiple videos with citations. Video subsumes both visual and audio modalities, making it a challenging testbed. We also validate that our evaluation generalizes independently to text, audio, and video.

\paragraph{Claim Decomposition} is a core component of factuality assessment. An important precursor is the Pyramid method \cite{nenkova-passonneau-2004-evaluating}, which abstracts summaries into Summary Content Units (SCUs)---minimal, semantically atomic propositions used as the basis for evaluation. SCUs introduced the principle that content should be assessed at the level of individual facts rather than full sentences, anticipating the subclaim decomposition central to modern factuality metrics. Building on this principle, FActScore \cite{min-etal-2023-factscore} and VeriScore \cite{song-etal-2024-veriscore} decompose predictions into subclaims and verify each one against a single text source, arguing that subclaims enable more straightforward factuality assessment than full sentences in virtue of their atomic propositional content. Subsequent work has attempted to refine the appropriate conception of \emph{atomicity} for subclaims \cite{gunjal-durrett-2024-molecular, wanner2024dndscoredecontextualizationdecompositionfactuality, wanner-etal-2024-closer} and has extended the core methodology to images \cite{jing-etal-2024-faithscore}. However, these metrics evaluate a prediction against a single source and do not assess citations, which requires verification across multiple sources. \mirage builds on this foundation but addresses a different setting: multi-source generation with citation attribution, where each subclaim must be factual and cited by the correct source.

\section{TextRAG Metrics}
\label{section:text-metrics}
Before introducing \frameworkname, we first revisit how RAG is assessed in text-only settings (TextRAG). We look at three prominent metrics---\acle \cite{gao-etal-2023-enabling}, \argue \cite{mayfield2024nuggets}, and \ragas \cite{es-etal-2024-ragas}---providing a foundation for RAG evaluation as well as the modifications to these metrics for handling multimodal sources. Implementation details for each are in  \autoref{append:text_rag}.

\paragraph{Preliminaries}
We define a \emph{document} as any single unit of evidence, regardless of modality (e.g., a text passage, image, audio clip, or video). We define a \emph{topic} as the subject matter of one or more \emph{documents}. The documents used in generation are the \emph{context}. A \emph{prediction} is text produced by a generation system, decomposed into \emph{subclaims}---atomic declarative statements that may or may not be true. A \emph{reference} is a collection of preexisting knowledge on a query topic, such as a knowledge base (e.g., Wikipedia), a set of nuggets \cite{Voorhees2004OverviewOT}, or a (human-written) reference text \cite{martin-etal-2026-wikivideo}. We prefix \emph{predicted} or \emph{reference} to denote the origin of derived objects (e.g., \emph{predicted subclaim}, \emph{reference subclaim}). A scoring function $\textsf{s}(p, h)$ measures support between a (premise, hypothesis) pair via entailment or LLM judgment.

\subsection{\acle}
\label{section:acle}

\acle \cite{gao-etal-2023-enabling} aims to capture three dimensions of predicted text quality: information correctness, citation quality, and fluency.\footnote{We do not explore fluency in this work, as it is not a common failure mode of modern LLMs.} 

\paragraph{Correctness} 
\acle introduces both recall and precision metrics to evaluate Correctness. Recall is a natural language inference- (NLI-)based judgment between reference subclaims and the predicted output; precision is exact match between predicted and reference answers. We do not implement precision, as exact match is unsuitable for long-form generation. We do not attempt to relax exact match in our implementation, as this would reduce to a metric like BERTScore, which we already include in our evaluation. Claim Recall requires no modification for multimodality, as verifying reference subclaims against a predicted text is independent of the modality of the sources.

\paragraph{Citation Quality}
\acle also uses recall and precision metrics to evaluate Citation Quality. Citation Quality Recall assesses whether all the sources cited by a sentence, jointly considered (i.e., concatenated), support the citing sentence. Citation Quality Precision penalizes irrelevant citations by assessing whether a sentence remains (jointly) supported by its cited sources when one of them is removed. Both precision and recall involve comparing a sentence against the concatenation of multiple text passages. In this work, we instead consider multiple multimodal documents.

Citation Quality has two limitations: (1)~sentences typically express multiple subclaims, making support harder to evaluate than with subclaims, and (2)~concatenating multiple documents is computationally infeasible  for long multimodal sources and unsupported by current training data.

\subsection{\argue}
\label{section:argue}

Similar to \acle, \argue \cite{mayfield2024nuggets} assesses generations for information coverage and for citation support. For the former, \argue adopts a \emph{nugget}-based evaluation, assessing what proportion of a set of question-answer pairs (\emph{nuggets}) about a target topic are attested by the generated text (Nugget Coverage). For the latter, \argue calculates the support between individual (sentence, citation) pairs (Sentence Support). We rely on the \autoargue implementation of \argue from \citet{walden-etal-2025-auto}. 

\paragraph{Nugget Coverage} is the proportion of nugget questions correctly answered by the predicted text, aggregated over sentences. This is similar to \acle's Claim Recall, but focused on recall of QA pairs rather than of subclaims. Since nugget coverage requires only these nuggets and the generated text for assessment, the only thing needed for our evaluation is to create the collection of nuggets.

\paragraph{Sentence Support} is the proportion of predicted sentences attested individually by each of their citations. In focusing on sentences rather than subclaims, Sentence Support runs into the same challenge as \acle's Citation Quality evaluation in having to simultaneously assess the support of multiple subclaims expressed by a single sentence.

\subsection{\ragas}
\label{section:ragas}

\ragas \cite{es-etal-2024-ragas} evaluates three dimensions of predicted text quality: faithfulness to the retrieved context (Faithfulness), relevance of the text to an information need (Answer Relevance), and relevance of the retrieved context to the information need (Context Relevance).

\begin{figure*}[t]
    \centering
    \includegraphics[width=\linewidth]{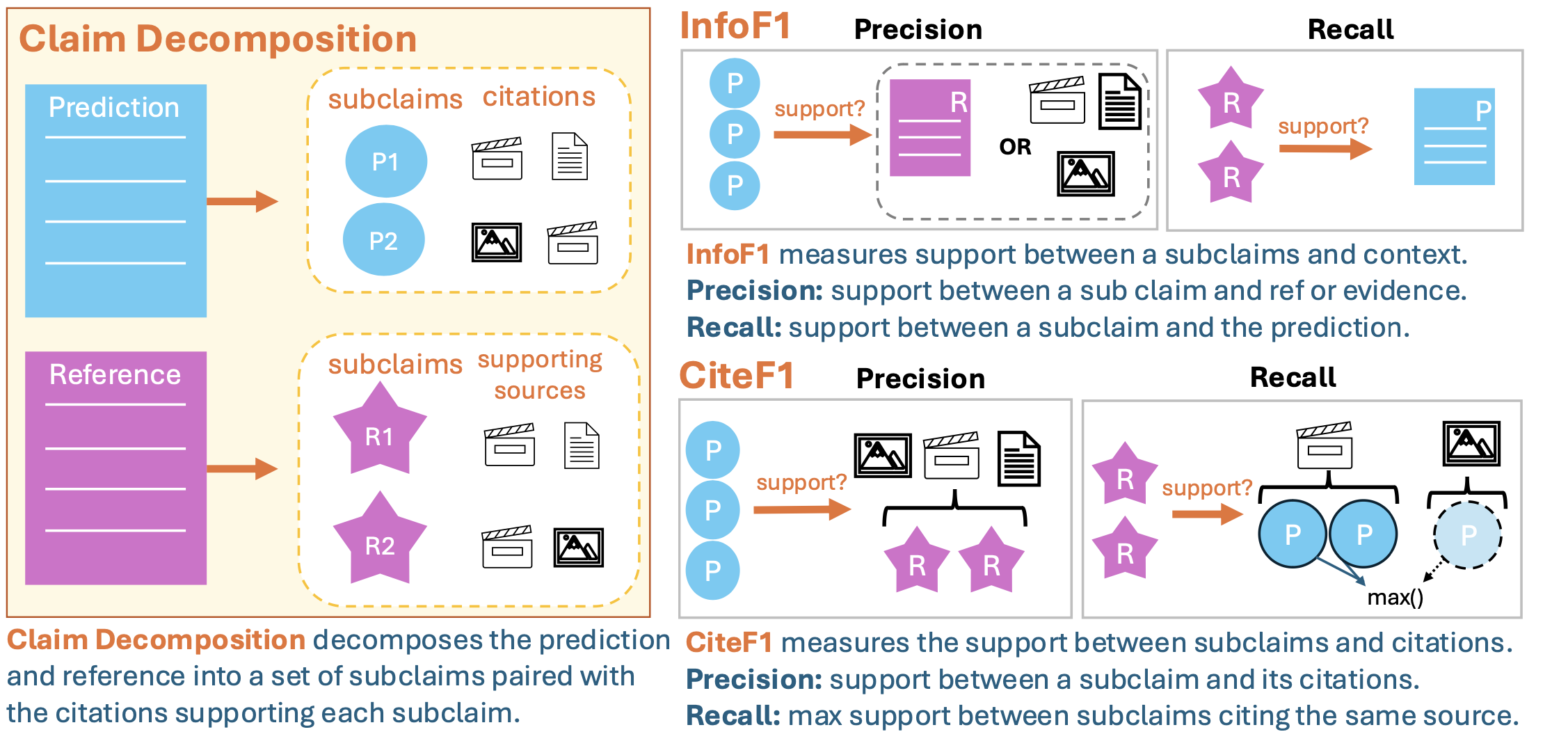}
    \vspace{-1.8em}
    \caption{\frameworkname decomposes generated text into subclaims to assess two dimensions of quality: \claimscore, which measures factuality and information coverage, and \citationscore, which measures citation precision and completeness. The framework enables consistent, claim-level evaluation across text, image, audio, and video modalities.}
    \label{fig:mirage}
    \vspace{-0.5em}
\end{figure*}

\paragraph{Faithfulness}
For Faithfulness, sentences from the predicted text are decomposed into ``statements'' (i.e., subclaims\footnote{\citet{es-etal-2024-ragas} create one or more statements from a sentence, which we interpret as subclaim decomposition.}) and are compared against all documents in the retrieved context to obtain a binary judgment of support. In our implementation, we apply the same claim decomposition method as we use in \mirage to obtain these statements. We evaluate statement support against the documents in context one-by-one, rather than jointly, owing to context limitations.

\paragraph{Answer Relevance}
For Answer Relevance, \ragas first uses an LLM to predict a query that could have yielded the predicted text as a response. Answer Relevance is then computed as the cosine similarity between the embeddings of the predicted and actual queries, given some embedding model.

\paragraph{Context Relevance} 
\ragas uses an LLM to extract sentences from the retrieved context that are judged to be essential to answering the query, computing the Context Relevance as $\frac{\text{extracted sentences}}{\text{total context sentences}}$. To adapt this for multimodal documents (videos, in our experiments), we extract query-relevant observations from each document and (separately) elicit a detailed VLM summary of the same document to approximate the denominator.

\subsection{Summary}
\label{section:summary}

The three metrics above share core limitations that hinder their applicability in multimodal domains: \acle and \argue assess information at the sentence level, making fact verification less precise. Additionally, \acle relies on concatenating multiple sources, which is infeasible with multiple long multimodal documents, and \ragas assumes an exhaustive extraction of observations from sources---an impracticable expectation in the multimodal domain. \mirage addresses these issues by decomposing predictions into subclaims across all evaluation dimensions, enabling more granular assessments of factuality and citation quality. Its two components, \claimscore and \citationscore, can be computed in both reference-based and reference-free settings, providing a scalable, claim-centric framework for evaluating multimodal RAG.

\section{\frameworkname}
\label{section:method}

\mirage (\autoref{fig:mirage}) is a subclaim-centric evaluation framework that evaluates two dimensions of predictions: the presented \textbf{information} and the use of \textbf{citations}. \claimscore measures both the proportion of predicted subclaims that are factual (precision) as well as the proportion of a set of reference subclaims that are supported by the predicted text~(recall). \citationscore measures whether subclaims are supported by their cited sources (precision) and whether reference subclaims attested by the prediction are attributed to a source that supports them~(recall). \autoref{append:metric_detail} provides further variants of \mirage for single-source inference and subclaim and source importance weighting.

\paragraph{Subclaim Decomposition}
A core step in \mirage is decomposing sentences into subclaims. We decompose predictions one sentence at a time using an LLM prompted to create a set of \emph{decontextualized} subclaims \cite{gunjal-durrett-2024-molecular, wanner2024dndscoredecontextualizationdecompositionfactuality}, ensuring that subclaims are atomic but independently verifiable. In our experiments, we use manually-authored reference subclaims, though automatic decomposition can be applied to other reference forms as well.

\paragraph{Reference Evaluations}
A key aspect of \mirage is the ability to evaluate in both reference-free (Collection) and reference-based (Reference) settings. In the Collection setting, subclaims are verified directly against the documents themselves. In the Reference setting, each document is mapped to a text-based proxy for its information content (e.g., a human-written article or set of manually authored subclaims). This makes Reference evaluation more computationally efficient, as it avoids multimodal inference, but introduces the limitation that human-authored references are inherently non-exhaustive: factual subclaims omitted from the reference will be penalized.

\subsection{\claimscore: Evaluating Information Quality}
We evaluate information quality using an F1 score over two metrics: \claimprec, which assesses subclaim factuality, and \claimrec, which assesses information coverage. We refer to the resulting metric as \claimscore.

\subsubsection{Information Factuality (\claimprec)}
We evaluate the factuality of a predicted text $P$ against a reference set of evidence documents $R=\{r_1,\ldots,r_n\}$. We sentence-split the predicted text and then decompose each predicted sentence into a set of subclaims ($C_P$). These subclaims are then scored for support against $R$ using a scoring function \textsf{s}, where \claimprec is defined as the proportion of predicted subclaims that are supported, weighted by the support strength given by $\textsf{s}$:
\begin{equation}
    \text{precision}(C_P, R) = \frac{1}{|C_{P}|}\sum_{c \in C_{P}} \textsf{s}(c, R)
\end{equation}

We propose two variations of \claimprec, Collection Precision and Reference Precision, which differ in how they construct $R$.

\paragraph{Collection Precision} defines $R$ as the set of relevant documents for the query topic. It is thus similar to \ragas Faithfulness, which evaluates support against the retrieved documents used as generation context, although our experiments rely on the gold set of relevant documents for each topic. In principle, $R$ could be a very large collection (e.g., a knowledge base), which would make it akin to a multimodal FActScore \cite{min-etal-2023-factscore}.

\paragraph{Reference Precision}
defines $R$ as the reference text or reference subclaims for the topic, evaluating support between predicted subclaims and the text proxy rather than the documents directly.

\subsubsection{Information Coverage (\claimrec)}
Information Coverage (\claimrec) is the proportion of subclaims in $R$ that are attested in $P$. It is exactly analogous to \claimprec, but with the roles of $R$ and $P$ reversed. If $R$ is given as a complete reference text, we first sentence-split $R$ and then decompose each sentence into a set of subclaims $C_R$.\footnote{If $R$ is given as a set of subclaims, we use these directly.}
\begin{equation}
    \text{recall}(C_R, P) = \frac{1}{|C_R|}\sum_{c\in C_R}\textsf{s}(c,P)
\end{equation}

\claimrec is the same as \acle Claim Recall, with both checking if subclaims in some reference are attested by the prediction.

A \emph{collection} vs.\ \emph{reference} distinction, analogous to the one for \claimprec, can be made here for \claimrec. In our experiments, we use the reference version of \claimrec, as the collection version imposes impractical computational costs and task complexity.

\subsection{\citationscore: Evaluating Citations}
\citationscore measures how effectively $P$ grounds information in cited sources. Like \claimscore (and unlike \acle and \argue), \citationscore works with subclaims. It is an F1 score over \citeprec---the proportion of subclaims supported by their cited sources---and \citerec---the proportion of subclaims that properly attribute their sources.

\subsubsection{Citation Support (\citeprec)}
\citeprec is the proportion of predicted subclaims that are supported by their cited documents, weighted by the strength of that support. We first decompose $P$ into a set of subclaims $C_P$ and create a mapping from each subclaim $c_i^P$ to the set of documents $D_i$ that are cited by the sentence from which $c_i^P$ was decomposed; i.e., $\{(c_i, D_i)\:\vert\: c_i^P \in C_P\}$. Given the support scoring function $\textsf{s}$, we compute:
\begin{equation}
    \text{precision}(C_P, D) = \frac{1}{|C_P|}\sum_{c_i^P \in C_P} \textsf{s}(c_i^P, D_i)
    \label{eq:cite_prec_main_text_3}
\end{equation}

\noindent As with \claimprec, \citeprec has \emph{collection} and \emph{reference} variants, which differ in the form of the cited documents used in evaluation.

\paragraph{Collection Precision} evaluates support between subclaims in $P$ and their associated cited documents directly---i.e., the raw text (or video) content. This is similar to \argue's Sentence Support; however, instead of evaluating support between a sentence and citation, we evaluate support between a \emph{subclaim} and a cited document.

\paragraph{Reference Precision} evaluates support between subclaims in $P$ and subclaims obtained from the associated cited documents, rather than the raw document content. In our experiments, we create a mapping from each cited document $d_i$ to the set of reference subclaims $C_{R,d_i}$ attested by $d_i$: $\{(d_i, C_{R,d_i})~|~P \text{ cites } d_i\}$, and evaluate support against $C_{R,d_i}$ rather than the document itself:
\begin{equation}
    \textsf{s}(c_i^P, C_{R,d_i})
\end{equation}

Prior metrics, \argue and \acle, evaluate full sentences against their citations, but this requires the sentence to be fully supported by \emph{each} document  (\argue) or the cited documents (\acle). This overpenalizes multi-premise sentences for containing un- or previously supported subclaims alongside supported ones, generally rewarding unnatural texts with short, single-subclaim sentences.

\subsubsection{Proper Attribution (\citerec)}
\citerec is the proportion of \emph{reference} subclaims ($C_R$) that are attested by $P$ via sentences that cite valid sources---i.e., sources that \emph{also} attest those subclaims. We first decompose the reference $R$ into subclaims and create a mapping from each subclaim $c_i^R$ to the set of documents $D_i$ in the collection that (individually) attest that subclaim: $\{(c_i^R,D_i)\:|\:c_i^R \in R\}$. Next, we create another mapping from each document $d_j$ that is cited in $P$ to the set $S_j$ of sentences in $P$ that cite $d_j$: $\{(d_j,S_j)\:\vert\:d_j \in P\}$. We then compose these two mappings to obtain the proportion of reference subclaims that are both attested in $P$ \emph{and} supported by a valid source citation:

\begin{equation}
\begin{split}
    &\text{recall}(C_R, P, D) = \\
    &\frac{1}{|C_R|}\sum_{c_i^R\in C_R} \max_{d_j \in D_i}\bigg(\max_{s_k \in S_j}~\textsf{s}(c_i^R, s_k)\bigg)
\end{split}
\end{equation}

\noindent We take the max support score over all documents $d_j \in D_i$ that attest the target subclaim ($c_i^R$)---and over all sentences $s_k$ in $P$ that cite $d_j$---because we care about the \emph{best} support that $P$ provides for each $c_i^R$ \emph{anywhere in the text}.

Unlike \claimrec, \citerec does not have the distinction between collection and reference settings, as \citerec relies on the set of \emph{reference} subclaims, ensuring that they are recalled with a valid source.

\begin{table}[t]
    \centering
    \setlength{\tabcolsep}{4pt}
    \begin{tabular}{lc|cccc}
    \toprule
        \textbf{Model} & \textbf{P} & \textbf{Text} & \textbf{Audio} & \textbf{Video} & \textbf{AudVid} \\
    \midrule
        AF & 7B & 50.9 & 60.0 & $-$ & $-$ \\
        Q2-A & 7B & 50.9 & 77.9 & $-$ & $-$ \\
    \midrule 
        Q2.5-V & 7B & 64.0 & $-$ & 22.1 & $-$ \\
        Q3.5 & 9B & 60.1 & $-$ & 76.3 & $-$ \\
        Q3.5 & 27B & \textbf{73.3} & $-$ & \textbf{79.6} & $-$ \\
        
    \midrule
        Q2.5-O & 7B & 65.3 & 81.9 & 50.5 & 45.2 \\
        \clue & 3B & 63.9 & \textbf{89.8} & 74.6 & \textbf{70.1} \\       
    \bottomrule
    \end{tabular}
    \caption{Accuracy of claim grounding methods. P: Parameters. AF: Audio Flamingo. Q: Qwen, A-Audio VL-Vision Language, O-Omni.}
    \label{tab:binary_nli}
    \vspace{-1em}
\end{table}

\section{Grounding Claims in Multimodal Data}
\label{section:grounding}
An important component of \mirage is the ability to ground claims in multimodal data. We evaluate claim grounding across four modality settings---Text (\unli \cite{chen-etal-2020-uncertain}; PeopleProfile \cite{walden2025groundedwikipediastudystructured}), Audio (\clotho \cite{drossos2020clotho, deshmukh2024audioentailmentassessingdeductive}; \wikivideo-Audio), Video, and Audiovisual (\wikivideo)---using audio-language \cite{goel2025audioflamingo3advancing, chu2024qwen2audiotechnicalreport}, vision-language \cite{bai2025qwen25vltechnicalreport, qwen35blog}, and omnimodal models \cite{xu2025qwen25omnitechnicalreport, zhang2026unifiedmultimodaluncertaininference}. Full details are in \autoref{append:exp_setup_details}.

In \autoref{tab:binary_nli}, we find that most methods struggle with claim verification. Modality-specific models perform decently in their respective modalities but lack the coverage needed for general-purpose multimodal evaluation, while omnimodal models offer broader coverage at varying quality. Both \clue \cite[omni;][]{zhang2026unifiedmultimodaluncertaininference} and Qwen3.5-27B \cite[vl;][]{qwen35blog} achieve strong performance on their respective modalities.

\section{Alignment With Human Judgments}
\label{section:human-alignment}
\paragraph{Evaluation Setup}
We evaluate \mirage on three datasets spanning text and audiovisual RAG. Together these let us bracket the difficulty spectrum of multimodal evaluation: RAGTIME \cite{lawrie2026overviewtrec2025ragtime}, a multilingual text-based report-generation task in the text-only setting for which ARGUE was designed; and \wikivideo \cite{martin-etal-2026-wikivideo} and \wikivideo-25 \cite{martin2026findingsmagmar2026shared}, two audiovisual (video) article-generation tasks that require grounding claims in long-form video. RAGTIME and \wikivideo-25 are shared tasks, for which we evaluate all predictions submitted to the task \cite{AMU-trec2025-papers-proc-1, WueRAG-trec2025-papers-proc-1, GenAIus-trec2025-papers-proc-6, HLTCOE-trec2025-papers-proc-2, IDACCS-trec2025-papers-proc-1, ncsu-las-trec2025-papers-proc-1, bhosale2026craftcriticrefinedadaptivekeyframe, chakraborty2026marquisthreestagepipelinevideo, yan2026traceevidencegroundingguidedmultivideo}. For \wikivideo, we generate predictions with three CAG \cite{martin-etal-2026-wikivideo} system variations (no videos, oracle, retrieved) across 10 topics and collect our own human judgments.

\begin{table}[]
    \centering
    \begin{tabular}{l|rrrr}
    \toprule
        & EQJ 1 & EQJ 2 & EQJ 3 & Avg \\
        \midrule
        ICJ 1   & $49.0$ & $61.3$ & $54.5$ & $54.9$ \\
        ICJ 2   & $65.3$ & $76.3$ & $70.8$ & $70.8$ \\
        ICJ 3   & $60.5$ & $81.2$ & $72.7$ & $71.4$ \\
        \midrule
        ICJ Avg & $58.3$ & $73.0$ & $66.0$ & $65.7$ \\
        \midrule
        GJ      & $18.2$ & $-5.5$ & $23.0$ & $11.9$ \\
    \bottomrule
    \end{tabular}
    \caption{Kendall's Tau agreement between human-annotated \mirage (ICJ[1-3]/GJ) and human-annotated quality judgments (EQJ[1-3]).}
    \label{tab:human_metric}
    \vspace{-1em}
\end{table}

\paragraph{Collection of Human Judgments}
We collect three types of judgments from human annotators.\footnote{See \autoref{append:human_judge} for instructions.}
\begin{enumerate}
\item \textbf{Extrinsic Quality Judgments (EQJs)}. For each prediction, annotators provide a holistic 1--5 Likert score based on the following attributes: factuality, adequacy, coherence, relevancy, and fluency.
\item \textbf{Intrinsic Claim Judgments (ICJs)}. Annotators are given the prediction, predicted subclaims, reference, and reference subclaims, and are asked to annotate whether or not a subclaim is supported by the other text. This directly mirrors \claimscore-Ref: the scores are what \claimscore would produce if a human performed the verification instead of a model.
\item \textbf{Grounding Judgments (GJs)}. Annotators are given all relevant videos for a topic and the predicted subclaims and judge whether or not each subclaim is supported by (grounded in) the videos. This directly mirrors \claimprec and \citeprec: the scores are what these metrics would produce with human verification.
\end{enumerate}
We collect these three sets of judgments for \wikivideo. For RAGTIME, we instead use the human annotations collected by the task organizers, which cover two dimensions: \emph{question attestation} (Information coverage)---whether a nugget question is answered (covered) by a predicted report---and \emph{sentence support} (citation support)---whether a sentence is supported by the document(s) it cites. For \wikivideo-25, we similarly use the human annotations collected by the task organizers, who provide a holistic 1--5 Likert score for each prediction, mirroring our EQJs. 

In all cases, we use Kendall's $\tau$ \cite{Kendall1938ANM} to assess system-level agreement between rankings induced by a given automatic metric and the appropriate set of human judgments.

\begin{table}[t]
\centering
    \begin{tabular}{l|rr}
    \toprule
        Metric Suite & Info Coverage & Cite Support \\
    \midrule
        ROUGE-L   & $-16.6$ & $45.9$ \\
        BERTScore & $29.8$ & $-11.5$ \\
        ARGUE     & $54.9$ & $66.3$ \\
        \mirage   & $\mathbf{77.7}$ & $\mathbf{79.2}$ \\
    \bottomrule
    \end{tabular}
\caption{Run-level Kendall's Tau between each metric suite and human judgments. For Auto-ARGUE, Information (Info) Coverage is Nugget Coverage and Citation (Cite) Support is Sentence Support. For \mirage, it's \claimrec and \citeprec, respectively, both in Collection.}
\label{tab:ragtime_agreement}
\vspace{-1em}
\end{table}

\begin{table*}[]
    \centering
    \begin{tabular}{ll|rrr|r|r|rr|rr}
    \toprule
         \multirow{2}{*}{Eval} & \multirow{2}{*}{Ano} & \multicolumn{2}{c|}{\claimprec} & \claimrec & \acle & \argue & \multicolumn{2}{c|}{\ragas} & \multirow{2}{*}{R-L} & \multirow{2}{*}{BS}\\
         & & \multicolumn{1}{r}{Ref.} & \multicolumn{1}{r|}{Col.} & \multicolumn{1}{r|}{Ref.} & \multicolumn{1}{r|}{Clm. Rec.} & \multicolumn{1}{r|}{Nugg. Cov.} & \multicolumn{1}{r}{Faith} & \multicolumn{1}{r|}{Ans. Rel.} & & \\
         \midrule
         \multirow{4}{*}{EQJ}
            & 1   & $50.8$ & $26.3$ & $59.2$ & $6.7$  & $-1.8$ & $15.2$ & $4.8$  & $55.8$ & $52.3$ \\
            & 2   & $38.2$ & $14.5$ & $84.2$ & $13.3$ & $10.0$ & $23.2$ & $23.3$ & $71.2$ & $81.2$ \\
            & 3   & $28.2$ & $21.5$ & $56.5$ & $14.8$ & $-8.5$ & $16.3$ & $-1.5$ & $41.2$ & $74.2$ \\
            & Avg & $39.1$ & $20.8$ & $66.6$ & $11.6$ & $-0.1$ & $18.2$ & $8.9$  & $56.1$ & $69.2$ \\
        \midrule
        \multirow{4}{*}{ICJ}
            & 1   & $84.8$ & $-10.0$ & $73.0$ & $34.8$ & $0.0$  & $11.8$ & $44.8$ & $66.7$ & $70.0$ \\
            & 2   & $66.7$ & $0.0$   & $72.7$ & $31.5$ & $13.3$ & $20.0$ & $33.3$ & $70.0$ & $86.7$ \\
            & 3   & $80.0$ & $0.0$   & $83.0$ & $18.5$ & $20.0$ & $26.7$ & $23.3$ & $68.2$ & $78.2$ \\
            & Avg & $77.2$ & $-3.3$  & $76.2$ & $28.3$ & $11.1$ & $19.5$ & $33.8$ & $68.3$ & $78.3$ \\
        \midrule
        GJ  & -   & $-6.7$ & $73.3$ & $-13.3$ & $-5.2$ & $-13.3$ & $3.0$ & $-26.7$ & $-11.5$ & $0.0$ \\
    \bottomrule
    \end{tabular}
    \caption{Kendall's Tau with all human judgments against the metrics that evaluate information. R-L: ROUGE-L, BS: BERTScore, EQJ: Extrinsic Quality Judgment, ICJ: Intrinsic Claim Judgment, GJ: Grounding Judgment.}
    \label{tab:wikivideo_info_agreement}
    \vspace{-1em}
\end{table*}

\begin{table}[t]
    \centering
    \setlength{\tabcolsep}{2pt}
    \begin{tabular}{@{}l|rrr|rrr@{}}
    \toprule
         \multirow{2}{*}{Eval} & \multicolumn{2}{c|}{\citeprec} & \citerec & \textsc{AC} & \textsc{ARG} & \textsc{RG} \\
         & \multicolumn{1}{c}{R} & \multicolumn{1}{c|}{C} & \multicolumn{1}{c|}{R} & \multicolumn{1}{r}{CQ} & \multicolumn{1}{r}{SS} & \multicolumn{1}{r}{CR} \\
         \midrule
        EQJ 1   & $3.3$  & $-4.8$  & $28.0$ & $-20.0$ & $-15.0$ & $32.7$ \\
        EQJ 2   & $31.5$ & $-19.7$ & $49.3$ & $10.0$  & $1.8$   & $23.2$ \\
        EQJ 3   & $40.0$ & $-6.7$  & $51.3$ & $8.2$   & $-1.8$  & $13.2$ \\
        EQJ A & $24.9$ & $-10.4$ & $42.9$ & $-0.6$  & $-5.0$  & $23.0$ \\
        \midrule
        GJ      & $-20.0$ & $26.7$ & $-6.7$ & $-8.5$ & $-3.3$ & $3.0$ \\
    \bottomrule
    \end{tabular}
    \caption{Kendall's Tau for all human judgments against citation metrics. AC: \acle, ARG: \argue, RG: \ragas. R: Reference, C: Collection.}
    \label{tab:wikivideo_cite_agreement}
    \vspace{-1em}
\end{table}

\subsection{Human-Annotated \mirage}
ICJs and GJs are human-computed instances of \mirage: they apply the same subclaim-level precision and recall, but with human annotators performing the verification step rather than a model. In \autoref{tab:human_metric}, we evaluate whether these human-computed \mirage scores agree with the extrinsic quality judgments (EQJs), which reflect holistic human assessments of output quality.

We find that \emph{when computed by humans, \claimscore strongly aligns with EQJs}. This demonstrates that with human-level verification performance, \mirage captures human quality preferences, validating the framework's design independently of any particular automatic verifier. We also observe that grounding judgments (GJs) show no-to-low agreement with EQJs across annotators. This suggests that groundedness is not well captured by holistic quality judgments alone. This result reinforces that \emph{metrics targeting quality or information coverage may not capture groundedness in sources unless they explicitly verify against those sources}.

\subsection{Automatic Evaluations}
We next evaluate the automatic implementations of \mirage and TextRAG metrics using Qwen3.5,\footnote{Further results with Qwen3.5 and \clue in \autoref{append:additional_alignment}.} reporting agreement with the human judgments on RAGTIME~(\autoref{tab:ragtime_agreement}), \wikivideo (\autoref{tab:wikivideo_info_agreement}, \autoref{tab:wikivideo_cite_agreement}), and \wikivideo-25 (\autoref{tab:all_magmar_scores_qwen35}).

\paragraph{\mirage best captures human preference.}
Across all three datasets and every judgment type, \mirage achieves the strongest and most consistent agreement with human judgments. On RAGTIME (\autoref{tab:ragtime_agreement}), \mirage attains the highest agreement on both information coverage and citation support, outperforming text-specific metrics for summarization and RAG. On \wikivideo, \claimscore (\autoref{tab:wikivideo_info_agreement}) aligns strongly with both EQJs and ICJs,\footnote{ICJs are scored against the reference, so a subclaim supported by the source collection but omitted from the reference counts as supported under Collection yet uncredited by the annotator, lowering agreement. See \autoref{append:refvcollect_prec}.} and \citationscore (\autoref{tab:wikivideo_cite_agreement}) aligns well with EQJs. Most notably, when verifying subclaims against the video content, \mirage is the only metric to achieve positive agreement with grounding judgments (GJs). Finally, \claimprec remains positive on a third dataset, \wikivideo-25 (\autoref{tab:all_magmar_scores_qwen35}). \emph{\mirage is thus not only the strongest metric on text, but the only one that extends to multimodal RAG---generalizing at no cost to text performance.}

\begin{table}
    \centering
    \setlength{\tabcolsep}{3pt}
    \begin{tabular}{ll|rrr|rr}
    \toprule
         \multirow{2}{*}{Eval} & \multirow{2}{*}{Ano} & \multicolumn{2}{c|}{\claimprec} & \claimrec & \multirow{2}{*}{R-L} & \multirow{2}{*}{BS} \\
         & & \multicolumn{1}{c}{R} & \multicolumn{1}{c|}{C} & \multicolumn{1}{c|}{R} & & \\
    \midrule
    \multirow{4}{*}{EQJ}
        & 1   & $29.4$ & $18.3$ & $-22.0$ & $-4.0$ & $9.0$  \\
        & 2   & $-7.3$ & $-18.3$ & $7.3$  & $-1.0$ & $7.0$  \\
        & 3   & $26.2$ & $15.0$ & $-44.9$ & $16.0$ & $15.0$ \\
        & Avg & $16.1$ & $5.0$  & $-19.9$ & $3.7$  & $10.3$ \\
    \bottomrule
    \end{tabular}
    \caption{Kendall's Tau on \wikivideo-25 between \mirage, ROUGE (R-L), and BERTScore (BS), and human judgments. R: Reference, C: Collection.}
    \label{tab:all_magmar_scores_qwen35}
    \vspace{-1em}
\end{table}

\paragraph{Evaluate at the subclaim level.}
Comparing \mirage's subclaim-level evaluation against the sentence-level approaches of \acle and \argue reveals a consistent pattern: evaluating at the sentence level reduces agreement with human judgments, regardless of modality. On RAGTIME (\autoref{tab:ragtime_agreement}), \mirage's subclaim-level \citeprec outperforms \argue Sentence Support, and its \claimrec outperforms \argue Nugget Coverage. On \wikivideo (\autoref{tab:wikivideo_cite_agreement}), \acle Citation Quality and \argue Sentence Support correlate more weakly with EQJs than their \mirage counterparts and have negative agreement with grounding judgments. This extends the findings of the subclaim literature in text \cite{min-etal-2023-factscore} to the multimodal setting. \emph{Atomic subclaims (not sentences) are the appropriate unit of evaluation across text and multimodal RAG alike.}

\paragraph{Surface-level metrics fail to generalize and ground.}
On \wikivideo, where high-quality human-written articles exist for each topic, ROUGE and BERTScore achieve reasonable agreement with quality (EQJ) and information (ICJ). However, they fail to capture grounding, producing near-zero or negative correlation with GJs, failing to verify information against its sources. This agreement does not generalize across tasks, where less comprehensive references exist. On RAGTIME (\autoref{tab:ragtime_agreement}), both metrics have negative agreement (ROUGE in information coverage and BERTScore with citation support) with human judgments, and on \wikivideo-25 (\autoref{tab:all_magmar_scores_qwen35}), their agreement with EQJs drops to near-zero. Thus, \emph{surface-level metrics are brittle proxies for quality, dependent on high-quality references and unable to assess grounding.}

\paragraph{Automatic verification remains a bottleneck.}
\mirage's framework design is validated by the strong agreement between human-computed scores and quality judgments (\autoref{tab:human_metric}) and its performance in text, where LLMs are widely accepted as competent verifiers \cite{min-etal-2023-factscore, song-etal-2024-veriscore}. Its automatic variants do not reach the same agreement with audiovisual sources, however, because the verifier must ground claims in video. As \autoref{section:grounding} shows, the best omnimodal verifier achieves only modest accuracy on audiovisual verification. The gap between human-annotated \mirage and its automatic counterpart thus emphasizes (1) that \mirage with human-level grounding can achieve strong agreement and (2) the need for future work in improving the claim verification against multimodal evidence.

\section{How to Evaluate Multimodal RAG}
\label{section:how-to}
We offer two recommendations for evaluating multimodal RAG. For quick diagnostics or compute-constrained settings, we recommend reference-based \claimscore and \citationscore as fast checks for factuality and grounding, verifying claims against a text proxy rather than the source documents. For comprehensive evaluation, we recommend collection-based \claimscore and \citationscore, which verify claims directly against the source documents and thereby avoid the limitations of surface-level and sentence-level metrics shown above. ROUGE and BERTScore remain reasonable proxies when high-quality references exist, and RAGAS Faithfulness and ALCE Claim Recall closely mirror components of \mirage. Evaluating answer relevance, as RAGAS Answer Relevance does (although not perfectly), also remains important. \textbf{For \mirage, we recommend reporting Precision and Recall alongside F1, as they are fundamental to the metric, while the F1 is derived.}

\section{Conclusion}
We introduce \mirage, a subclaim-centric framework for evaluating multimodal RAG. Through \claimscore and \citationscore, \mirage provides consistent, subclaim-level assessment of factuality, information coverage, and citation quality across text, image, audio, and video. We show that \mirage outperforms text-centric and sentence-level metrics on text while remaining the only metric to generalize to multimodal sources, aligning with human quality judgments where the others do not. As future work continues to improve omnimodal verification, \mirage will enable scalable and reliable evaluation for multimodal generation.

\section*{Limitations}
\paragraph{Evaluating Information Refusal}
\mirage aims to evaluate the information presented in a prediction. However, there is one aspect we don't capture: refusal (e.g., ``There is no information on the location of the fire in the videos.''). Determining that a subclaim is not supported by or contained in a large multimodal collection is a challenging task, requiring either strong a priori knowledge of the collection or perfect retrieval and grounding within it. Additionally, claims refusing to answer are out-of-distribution for our entailment judges. 

\paragraph{Multi-Video Inference Constraints}
As mentioned, the inability to provide multi-video inference limits the capabilities of each method presented. Extreme reduction of video frame rates to enable multi-video inference limits the performance of the VLM verifier and performing only single video judgments limits the ability to assess any cross-source inferences made by the generation system. Training VLMs for multi-video inference and optimizing multimodal inference are much needed future work for more holistic (multi-video) and scalable (optimization) evaluation.

\bibliography{anthology, custom}

\appendix

\section{Experimental Setup Details}
\label{append:exp_setup_details}

In this section, we outline in more detail the experiments in \autoref{section:grounding} and \autoref{section:human-alignment}.

\subsection{Claim Grounding Setup}
In \autoref{section:grounding}, we explored how to ground claims in multimodal data. In this section, we breakdown how we setup the data, ran prediction models, and the model hyperparameters.

\subsubsection{Data}
In our evaluation, we report the claim verification performance on four modality combinations: text, audio, video, and audiovisual. 

\paragraph{Text Only.}
For text data, we evaluate the performance of claim verification on \unli \cite{chen-etal-2020-uncertain} and PeopleProfile \cite{walden2025groundedwikipediastudystructured}. \unli is a standard natural language inference task between a premise and hypothesis in text. PeopleProfile is a dataset from an exploration into the groundedness of a Wikipedia article in the cited sources. These annotations contain scalar support judgments ($-1.0$ to $1.0$) from GPT-4o-mini between a subclaim and source article. This data, although synthetic, provides a verification task more salient to our evaluation setting. For both datasets, we binarize the labels, treating scores above 0.5 as supported and below 0.5 as not supported.

\paragraph{Audio Only.}
For the audio data, we evaluate the performance of claim verification on \clotho \cite{drossos2020clotho} and the \wikivideo subclaims that are supported by audio only. The \clotho dataset has been modified by \citet{deshmukh2024audioentailmentassessingdeductive} as an audio-NLI task between an audio premise and text hypothesis with contradictory, neutral, and entailment values. We take only the contradictory (not supported) and entailment (supported) instances from the data. We leave \wikivideo audio claims unmodified.

\paragraph{Video only.}
For video only, we take the \wikivideo subclaims that are supported by video and/or OCR only. Other Video-NLI data exists in the literature \cite{liu2020violinlargescaledatasetvideoandlanguage}, however, the underlying videos are no longer distributed. We use the \wikivideo video claims unmodified.

\paragraph{Audiovisual.}
For audiovisual, we again take the \wikivideo subclaims, but this time we take any combination of modalities, leaving the single-modality support for their respective modalities. We use the \wikivideo-audiovisual claims unmodified.

In the \wikivideo data, there are subclaims that are unsupported by any modality. We mix these in (providing a balanced 50/50 label distribution) for the unsupported judgments with unique, randomly selected claims for each modality instance.

\subsubsection{Models}
We evaluate models spanning four capability profiles: audio-only (Audio Flamingo 3 (AF) \cite{goel2025audioflamingo3advancing}, Qwen2-Audio (Q2-A) \cite{chu2024qwen2audiotechnicalreport}), language-only (Qwen3 (Q3) \cite{yang2025qwen3technicalreport}), vision-language (Qwen2.5-VL (Q2.5-VL) \cite{bai2025qwen25vltechnicalreport}, Qwen3.5 (Q3.5) \cite{qwen35blog}), and omnimodal (Qwen2.5-Omni (Q2.5-O) \cite{xu2025qwen25omnitechnicalreport}, \clue \cite{zhang2026unifiedmultimodaluncertaininference}).

\subsubsection{Hyperparameters}
The hyperparameters used for each model are listed in \autoref{tab:hyperparameters}.

\begin{table}[]
\centering
\begin{tabular}{lr}
\toprule
\textbf{Hyperparameter} & \textbf{Value} \\ \midrule
GPU Memory Utilization & $0.9$ \\
Max New Tokens & $4{,}096$ \\
Batch Size & $4$ \\
FPS & $0.5$ \\
Max Pixels & $256 \times 256$ \\
Min Pixels & $256 \times 256$ \\
Temperature & $0.7$ \\
Top-$p$ & $0.95$ \\ \bottomrule
\end{tabular}
\caption{Model hyperparameters.}
\label{tab:hyperparameters}
\end{table}

\subsection{Human Alignment}
In \autoref{section:human-alignment}, we explored how human and automatic metrics correlate when evaluating predictions. In this section, we breakdown how we set up and generated predictions. 

\paragraph{\wikivideo-25 (MAGMaR Shared Task)}
The MAGMaR shared task releases its submissions and a set of human evaluations. These human evaluations are a scalar score for the quality of a prediction and an overall ``best'' system selection. We use their scalar score, as we also use a scalar score for our \wikivideo annotations.

\paragraph{\wikivideo}
For \wikivideo, we generate predictions in three CAG \cite{martin-etal-2026-wikivideo} settings across 10 topics. 
\begin{enumerate}
    \item \textbf{LLM Only.} We generate a prediction without any information from the videos. Then, every sentence is used to retrieve one video as a citation from the oracle video set using OmniEmbed \cite{ma2025tevatron20unifieddocument}.
    \item \textbf{Oracle Video.} We generate a prediction with the oracle video set. For sentences without citations, we retrieve a video from the oracle video collection using Video-ColBERT \cite{reddy2025videocolbertcontextualizedlateinteraction}.
    \item \textbf{RAG.} We first retrieve 5 videos from MultiVENT2.0 \cite{kriz2025multivent20massivemultilingual} with MMMORRF \cite{samuel-etal-2025-mmmorrf}. Then we generate a prediction with the set of videos in the same way as oracle generation. 
\end{enumerate}

\section{\mirage++}
\label{append:metric_detail}

In the main text, we present \claimscore and \citationscore in their standard formulations. Here, we provide variations for single-source inference and weighted evaluation, and discuss the tradeoffs between reference and collection precision.

\subsection{\claimscore}

\subsubsection{Single-Source Precision}
Single-source precision is the more computationally feasible counterpart to the multi-source formulation. Instead of scoring a subclaim against all evidence, we compare the subclaim individually against each item in the evidence and take the maximum score:
\begin{equation}
    \text{precision}(C_P, R) = \frac{1}{|C_{P}|}\sum_{c \in C_{P}} \max_{d\in R}\textsf{s}(c, d)
\end{equation}
where $d$ denotes any source of information. While single-source precision is more computationally feasible, it cannot evaluate any cross-source inferences made by a prediction system.

\subsubsection{Weighted Variations}
We use a notion of \emph{information importance} to weight both precision and recall. For \textbf{precision}, information importance should be defined by the author of the prediction. For example, a human could annotate a ranked list of subclaims from most to least important, or a calibrated model \cite[e.g.,][]{jiang2025conformallinguisticcalibrationtradingoff} could express confidence in the predicted subclaims as their importance. For \textbf{recall}, information importance should be defined by the evaluation data or the consumer of the information---in practice, annotating each reference subclaim for its importance to the queried information need.

We define $I_i$ as the importance for a subclaim $c_i$. Weighted precision is:
\begin{equation}
    \text{precision}(C_P, R) = \frac{1}{|C_{P}|}\sum_{c_i \in C_{P}} \textsf{s}(c_i, R) \cdot I_i
\end{equation}
For single-source inference:
\begin{equation}
    \text{precision}(C_P, R) = \frac{1}{|C_{P}|}\sum_{c_i \in C_{P}} \max_{d\in R}\textsf{s}(c_i, d) \cdot I_i
\end{equation}
Weighted recall is:
\begin{equation}
    \text{recall}(C_R, P) = \frac{1}{|C_R|}\sum_{c_i\in C_R}\textsf{s}(c_i, P) \cdot I_i
\end{equation}

\subsubsection{Implementation Details}
We support two verification backends. For the \clue verifier, we use the scalar prompts (Figures~\ref{prompt:scalar_audio},~\ref{prompt:scalar_vision},~\ref{prompt:scalar_text}) which return a continuous support score between 0 and 1, selecting the appropriate prompt based on document modality. With the scalar predictions we use a 0.5 threshold for supported/not supported. For the zero-shot VLM verifier, we use the binary yes/no prompts for video (\autoref{prompt:zs_prompt_infof1_video}) and text (\autoref{prompt:zs_prompt_infof1_text}) documents, and the subclaim-level prompt (\autoref{prompt:zs_prompt_citef1_llm}) for Reference Precision. In both cases, Collection Precision verifies each predicted subclaim against each document individually using single-source precision and takes the maximum score. Information Coverage uses the same prompts with the roles of prediction and reference reversed.

\subsection{\citationscore}

\subsubsection{Single-Source Precision}
As with \claimscore, we reformulate citation precision for single-source inference. Instead of scoring a subclaim against all of its cited sources simultaneously, we compare the subclaim against each cited source individually and take the maximum:
\begin{equation}
    \text{precision}(C_P, D) = \frac{1}{|C_P|}\sum_{c_i \in C_P} \max_{d \in D_i} \textsf{s}(c_i, d)
\end{equation}
The set of documents $D_i$ can either be the documents associated with the first mention of $c_i$, like \argue, or any mention of $c_i$ depending on the desired evaluation setting. Subclaim de-duplication is recommended as shown by \citet{jiang2024corerobustfactualprecision}.

\subsubsection{Weighted Variations}
For \citationscore, we use a notion of \emph{source quality} to weight both precision and recall. The quality of a source could be annotated in the evaluation data for two cases: (1) differentiating between counterfactual sources (e.g., AIGC or misinformation), and (2) human preferences. For example, a human might prefer that a system cite the most informative sources, so that fewer need to be consulted to validate a prediction. Alternatively, they might prefer a particular type of source (e.g., news content over social media).

We define $Q_i$ as the quality weight for a source. Weighted citation precision is:
\begin{equation}
    \text{precision}(C_P, D) = \frac{1}{|C_P|}\sum_{c_i \in C_P} \textsf{s}(c_i, D_i) \cdot Q_i
\end{equation}
For single-source inference:
\begin{equation}
    \text{precision}(C_P, D) = \frac{1}{|C_P|}\sum_{c_i \in C_P} \max_{d \in D_i} \textsf{s}(c_i, d) \cdot Q_d
\end{equation}
Weighted citation recall is:
\begin{equation}
\begin{split}
    &\text{recall}(C_R, P, D) = \\
    &\frac{1}{|C_R|}\sum_{c_i^R\in C_R} \max_{d_j \in D_i}\bigg(\max_{s_k \in S_j}~\textsf{s}(c_i^R, s_k) \cdot Q_j\bigg)
\end{split}
\end{equation}

\subsubsection{Implementation Details}
As with \claimscore, we support both \clue and zero-shot VLM verification backends, using the same scalar prompts (\autoref{prompt:scalar_audio},~\autoref{prompt:scalar_vision},~\autoref{prompt:scalar_text}) and binary prompts (\autoref{prompt:zs_prompt_infof1_video},~\autoref{prompt:zs_prompt_infof1_text},~\autoref{prompt:zs_prompt_citef1_llm}) respectively.

\subsection{Reference vs.\ Collection Precision}
\label{append:refvcollect_prec}
We discuss the tradeoffs between reference and collection precision in \claimscore and \citationscore. The primary motivation for reference-based evaluation in multimodal RAG is computational efficiency, but each variant introduces distinct side effects.

\paragraph{\claimscore.}
Reference precision evaluates predicted subclaims against the reference instead of against the documents related to the topic (collection precision). However, human-annotated references are inherently non-exhaustive, leading to the artifact that reference precision penalizes factual subclaims deemed not salient to the topic. This may be a limitation if the goal is to evaluate \emph{only} factuality: without judging subclaims against the information sources, the evaluation cannot capture the factuality of all claims in the prediction. We do not view this as a limitation for general evaluation, because it is important that information be relevant to the query and information need.

\paragraph{\citationscore.}
Reference precision for \citationscore evaluates predicted subclaims against the reference subclaims grounded in the same cited document. Like \claimscore, human-annotated references are non-exhaustive, so reference precision penalizes factual subclaims not deemed salient to the source topic. Unlike \claimscore, this \textbf{is a limitation} of reference precision: \citationscore is designed to evaluate grounding in the cited sources, so it should not matter whether the information is relevant to the topic---only that it is supported by the attributed sources.

\section{Additional Human Alignment Results}
\label{append:additional_alignment}
In \autoref{tab:ragtime_metric_combined}, \autoref{tab:wikivideo_metric_combined}, and \autoref{tab:magmar_metric_combined}, we report additional human alignment results with \clue and Qwen3.5.

\section{TextRAG Implementations}
\label{append:text_rag}
\subsection{\acle}
In \autoref{section:acle}, we introduce \acle. In this section, we outline the key implementation details of this metric. 

\paragraph{Correctness}
In our work, we only evaluate Claim Recall from \acle Correctness. We do not adapt exact match against the reference because this is not suitable for long-form generation. When evaluating open-ended RAG on ELI5 \cite{fan-etal-2019-eli5}, \citet{gao-etal-2023-enabling} present Claim Recall, an NLI-based metric between subclaims in the reference (gold answers) and the prediction (model output). This adapts to our setting with no modification and we use the scoring function from \claimrec.

\paragraph{Citation Quality}
\citet{gao-etal-2023-enabling} evaluate citation quality with an F1 metric of citation recall and precision. When evaluating citations, \citet{gao-etal-2023-enabling} compare a sentence from the prediction to the concatenation of cited sources. This poses two main issues in evaluation: (1) from a factuality perspective, evaluating at the sentence level requires verifying multiple subclaims in a single inference step \cite{min-etal-2023-factscore}, and (2) from a multimodal perspective, concatenating multiple videos raises issues with computational constraints and the non-existence of multiple video training data for VLMs \cite{li2025videochatflashhierarchicalcompressionlongcontext, martin-etal-2026-wikivideo}.

To faithfully implement citation quality, we wrap verification in a loop which downsamples the framerate of the video(s) until they fit on GPU memory. We first attempt to verify a (subclaim, videos) pair at 1 fps per video. If an out-of-memory (OOM) exception is thrown, the videos are downsampled by reducing the framerate of all videos by half ($\frac{\text{fps}}{2}$) until 10 iterations of downsampling or until no OOM is reached. We never reach the 10 iterations on the videos in \wikivideo, but allow for setting downsampling iterations to be defined by the evaluator. To verify the sentences we use the prompt in \autoref{prompt:zs_prompt_citation_recall_alce}.

\begin{figure*}[t]
\noindent\fbox{\parbox{.98\textwidth}{\small\ttfamily
\textbf{System Prompt:} You are an expert at verifying information. You will be given a set of videos and a sentence. Your task is to determine if the sentence is fully supported by the videos. You will output <response>yes</response> if the sentence is fully supported by the videos, or <response>no</response> if the sentence is not fully supported by the videos.

\textbf{User Prompt:} [VIDEOS\_HERE] Sentence: [PUT\_SENTENCE\_HERE] Is the sentence fully supported by the videos? Only respond with <response>yes</response> or <response>no</response>.
}}
\caption{Zero-shot prompt for \acle Citation Quality.}
\label{prompt:zs_prompt_citation_recall_alce}
\end{figure*}

\subsection{\argue}

\argue \citep{mayfield2024nuggets} is a RAG evaluation framework developed for \emph{report generation}---the task of producing a long-form, citation-attributed response to a complex user query. Under \argue, reports accumulate a series of rewards and penalties based on sentence-level judgments about whether each sentence is supported by its associated citations and whether it correctly answers any of a predetermined set of questions (or \emph{nuggets}) associated with the report topic. \argue also handles various nuances related to these two dimensions, including whether a sentence \emph{reiterates} information stated previously in the report (and thus does not require a citation) or whether a statement correctly asserts that some nugget question is unanswerable from the underlying collection. Our implementation of this metric is based largely on \autoargue \cite{walden-etal-2025-auto}.

\paragraph{Nuggets}
To evaluate predictions, \argue has a knowledge base of question-answer pairs on the topic called nuggets. None of our evaluation data contains nuggets, so we first create a set of nuggets to use in evaluation. To do this, we convert our reference subclaims to nuggets by generating questions with an LLM \cite[Qwen3;][]{yang2025qwen3technicalreport} and then pairing the questions and claim as a question-answer pair.

\paragraph{Sentence Support} is the proportion of predicted sentences attested by each of their citations. The open-source implementation of \argue \cite{walden-etal-2025-auto} we use implements this for a single citation at a time, which we follow. The only modification we make to this code is changing their verification of the sentence against a text document with the verification of the sentence against a video. 

\paragraph{Nugget Coverage} is the proportion of nugget questions correctly answered by the report, aggregated over sentences. This is similar to \acle Claim Recall or our \claimscore Recall, but instead recalling QA pairs. Like \acle, this score requires no modification for multimodal RAG evaluation. In our implementation of this, we do not use our scoring function with \clue. Instead, we use the LLM critic from the implementation. 

\subsection{\ragas}

\ragas \cite{es-etal-2024-ragas} is a RAG evaluation framework developed for the automatic evaluation of retrieval-augmented generation---the task of producing long-form, citation-attributed responses to a user query. \ragas consists of three main components: Faithfulness, Answer Relevance, and Context Relevance.

\paragraph{Faithfulness} evaluates statements from a prediction against the context used in generation. We interpret these statements as subclaims to resolve the factuality issues of \argue and \acle. Additionally, like Citation Quality from \acle, this experiences the same issue of computational infeasibility in multi-source inference. However, unlike \acle, Faithfulness is not designed such that it is necessary to perform inference on multiple sources. We modify this to verify one subclaim in one document at a time, but only using the videos used in generation and not all relevant videos. We use the same scoring function as \claimprec/\citeprec.

\paragraph{Answer Relevance} evaluates the embedding similarity between the query used during prediction and an LLM-generated query based on the prediction. This metric remains unchanged from \citet{es-etal-2024-ragas} because the generation and comparison of potential queries is text-based regardless of the information source modality. We compute this as the cosine similarity between the embeddings of the predicted and actual queries, using the Qwen3 Embedding model \cite{zhang2025qwen3embeddingadvancingtext}.

\paragraph{Context Relevance} extracts sentences from the context based on the query and computes relevance as $\frac{\text{num extracted sentences}}{\text{total number of sentences}}$. We attempt to implement this with a VLM, extracting all information relevant to the query (numerator) and extracting all information from the video (denominator). However, extracting all information from the video is challenging, as the number of potential claims is near infinite. We instead elicit a detailed summary from a VLM for the denominator. We extract information related to the query with \autoref{prompt:ragas_context_relevance_num} (numerator) and detailed video information with \autoref{prompt:ragas_context_relevance_denom} (denominator).

In the \ragas implementation, \citet{es-etal-2024-ragas} give the LLM the ability to back off and say that a source has ``insufficient information.'' We notice in our implementation that \wikivideo is challenging for VLMs to make high-level inferences about and thus leads to a large number of videos deemed ``insufficient information,'' even though humans have annotated and grounded relevant claims to the query topic in those same videos.

\begin{figure*}[t]
\noindent\fbox{\parbox{.98\textwidth}{\small\ttfamily
Please extract relevant sentences from the provided context that can potentially help answer the following question. If no relevant sentences are found, or if you believe the question cannot be answered from the given context, return the phrase `Insufficient Information'. The question is: [PUT\_QUESTION\_HERE]
}}
\caption{Zero-shot prompt for extracting query-relevant information used in the numerator of \ragas Context Relevance.}
\label{prompt:ragas_context_relevance_num}
\end{figure*}

\begin{figure*}[t]
\noindent\fbox{\parbox{.98\textwidth}{\small\ttfamily
Describe the video in detail and extract all the information possible from it. This includes transcribing any on-screen text (OCR) and describing any visual information beyond the summary.
}}
\caption{Zero-shot prompt for extracting the detailed summary used in the denominator of \ragas Context Relevance.}
\label{prompt:ragas_context_relevance_denom}
\end{figure*}

\section{Collection of Human Judgments}
\label{append:human_judge}

We collect three human judgments.

\begin{enumerate}
    \item \textbf{Extrinsic Quality Judgments (EQJs)}. Human annotators are given information about the topic and are asked to give a 1-5 likert score considering the following attributes (in order of importance): factuality, adequacy, coherence, relevancy, and fluency. 
    \item \textbf{Intrinsic Claim Judgments (ICJs)}. Human annotators are given the prediction, predicted subclaims, reference, and reference subclaims, and are asked to annotate whether or not a subclaim is supported by the other text. This directly mirrors \claimscore-Ref: the scores are what \claimscore would produce if a human performed the verification instead of a model. 
    \item \textbf{Grounding Judgments (GJs)}. Human annotators are given all the relevant videos for a topic and the predicted subclaims and asked to annotate whether or not the subclaims are supported by (grounded in) the videos. \claimprec and \citeprec: the scores are what they would produce with human verification.
\end{enumerate}

For each grounding judgment, we calculate the ranking between the three prediction settings and we use the scipystats \cite{2020SciPy-NMeth} implementation of Kendall's Tau \cite{Kendall1938ANM}.

\subsection{Extrinsic Quality Judgments}
\label{subappend:likert}
In \autoref{prompt:human_likert_instructions}, we provide the annotation instructions for the extrinsic quality judgments. For these judgments, we recruit three annotators who are native/fluent English speakers and provide them each the 10 evaluation instances and annotation instructions for the task. These annotations are redundant. We adapt these judgments from \cite{gantt-etal-2024-event}.

\subsection{Intrinsic Claim Judgments}
\label{subappend:intrinsic}
In \autoref{prompt:human_metric_instructions}, we provide the annotation instructions for intrinsic quality judgments. For these judgments, we recruit three annotators who are native/fluent English speakers and provide them each the 10 evaluation instances and annotation instructions for the task. These annotations are redundant.

\subsection{Grounding Judgments}
\label{subappend:grounding}
We use the same annotation instructions from \citet{martin-etal-2026-wikivideo} and refer the reader to their work for the annotation instructions. For these judgments, we recruit 5 annotators who are native/fluent English speakers and provide them the annotation instructions and the annotation data. These annotations are not redundant.

\subsection{Human \frameworkname Granular Alignment}
In \autoref{tab:human_metric_all} we provide the alignment between each component (Precision, Recall) of human annotated \claimscore and extrinsic quality judgments.

\section{Acknowledgment of AI}
AI assistants were used in this project for coding and to edit text for fluency and typos. 

\begin{figure*}[t]
\noindent\fbox{\parbox{.98\textwidth}{\small\ttfamily
\textbf{System Prompt:} You are an expert in evaluating and verifying claims. You will be given a video, a claim, and the context the claim came from. Your task is to determine if the claim is supported by the video. You will output <response>yes</response> if the claim is supported by the video, or <response>no</response> if the claim is not supported by the video.

\textbf{User Prompt:} [VIDEO\_HERE] Here is the context the claim came from: <claim\_context> [PUT\_CONTEXT\_HERE] </claim\_context>. Here is the claim: <claim> [PUT\_CLAIM\_HERE] </claim>. Only respond with <response>yes</response> or <response>no</response>. Is the claim: [PUT\_CLAIM\_HERE], supported by the video?
}}
\caption{Zero-shot prompt for claim verification in video.}
\label{prompt:zs_prompt_infof1_video}
\end{figure*}

\begin{figure*}[t]
\noindent\fbox{\parbox{.98\textwidth}{\small\ttfamily
\textbf{System Prompt:} You are an expert in evaluating and verifying claims. You will be given a passage of text, a claim, and the context the claim came from. Your task is to determine if the claim is supported by the passage of text. You will output <response>yes</response> if the claim is supported by the passage, or <response>no</response> if the claim is not supported by the passage.

\textbf{User Prompt:} Here is the passage: <verification\_context> [PUT\_VERIFICATION\_CONTEXT\_HERE] </verification\_context>. Here is the context the claim came from: <claim\_context> [PUT\_CONTEXT\_HERE] </claim\_context>. Here is the claim: <claim> [PUT\_CLAIM\_HERE] </claim> Only respond with <response>yes</response> or <response>no</response>. Is the claim: [PUT\_CLAIM\_HERE], supported by the passage?
}}
\caption{Zero-shot prompt for claim verification in text.}
\label{prompt:zs_prompt_infof1_text}
\end{figure*}

\begin{figure*}[t]
\noindent\fbox{\parbox{.98\textwidth}{\small\ttfamily
\textbf{System Prompt:} You are an expert in evaluating and verifying claims. You will be given a claim, the context the claim came from, and a list of claims to verify the claim against. Your task is to determine if the claim is supported by the list of claims. You will output <response>yes</response> if the claim is supported by the list of claims, or <response>no</response> if the claim is not supported by the list of claims.

\textbf{User Prompt:} Here is the list of claims to verify against: <verification\_context> [PUT\_VERIFICATION\_CONTEXT\_HERE] </verification\_context>. Here is the context the claim came from: <claim\_context> [PUT\_CONTEXT\_HERE] </claim\_context>. Here is the claim: <claim> [PUT\_CLAIM\_HERE] </claim> Only respond with <response>yes</response> or <response>no</response>. Is the claim: [PUT\_CLAIM\_HERE], supported by the list of claims to verify against?
}}
\caption{Zero-shot prompt for claim verification in citation (text).}
\label{prompt:zs_prompt_citef1_llm}
\end{figure*}

\begin{figure*}[t]
\noindent\fbox{\parbox{.98\textwidth}{\small\ttfamily
To help you make more accurate and consistent judgments, here is an expanded explanation of how to interpret and assign support percentages for audio-based evidence. These examples cover a range of real-world cases you may encounter in this annotation task.

100\% - Full and unambiguous support: The audio clearly and directly contains the exact event, statement, or content described in the claim. There is no need for guessing or interpretation --- the claim is fully verified by the audio.

80-100\% - Almost complete support: The main content of the claim is clearly supported by the audio, though there may be minor ambiguity in speaker identity, context, or completeness (e.g., partial recording, mild noise). Overall, the claim is strongly supported.

60-80\% - Strong partial support: The audio strongly suggests that the claim is true, but some key details may be missing, unclear, or ambiguous --- such as incomplete phrases, background noise, or partial conversations. The evidence is strong but not definitive.

40-60\% - Moderate partial support: There is some alignment between the audio and the claim, but large portions are missing, unclear, or open to interpretation. While the recording points in the same general direction as the claim, it lacks clarity or completeness for confident verification.

20-40\% - Minimal weak support: There are small verbal cues or contextual hints that could relate to the claim, but they are insufficient to provide confidence in its truth.

0-20\% - Very weak or speculative support: There may be the slightest indirect reference (such as a related topic or similar voice), but nothing concrete that verifies the claim.

0\% - No support or contradiction: The audio does not relate to the claim at all, or it directly contradicts it.

Based on the provided audio and text, evaluate the probability that the text statement is true. Your answer must be a decimal number between 0 and 1, and you must strictly follow the format below:

<answer>probability\_value</answer>

Where probability\_value is the result you calculate. The text to evaluate is: \{text\}
}}
\caption{Zero-shot scalar prompt for audio examples.}
\label{prompt:scalar_audio}
\end{figure*}

\begin{figure*}[t]
\noindent\fbox{\parbox{.98\textwidth}{\small\ttfamily
To help you make more accurate and consistent judgments, here is an expanded explanation of how to interpret and assign support percentages. These examples are designed to cover a range of real-world cases you may encounter in the annotation task.

100\% - Full and unambiguous support: The video clearly shows the exact event described in the claim. There is no need for guessing or interpretation.

80-100\% - Almost complete support: The main content in the claim is shown, but there may be minor ambiguity in location, identity, or completeness. The overall claims are supported by the video.

60-80\% - Strong partial support: The video strongly suggests the claim is true, but some critical details may be missing, obscured, or ambiguous, limiting the ability to confirm the claim with certainty. The video gives strong but not definitive support.

40-60\% - Moderate partial support: There is some alignment with the claim, but large portions are either missing, unclear, or open to interpretation. While the footage may point in the same general direction as the claim, it lacks the clarity or completeness needed for confident verification.

20-40\% - Minimal weak support: There are small visual or audio cues that could hint at the claim, but they are insufficient to be confident.

0-20\% - Very weak or speculative support: There may be the slightest indirect reference, such as a related object or setting, but nothing concrete happens.

0\% - No support or contradiction: The video does not relate to the claim at all, or it directly shows something opposite.

Based on the provided video and text, evaluate the probability that the text is true. Your answer must be a decimal number between 0 and 1, and you must strictly follow the format below:

<answer>probability\_value</answer>

Where probability\_value is the result you calculate. The text to evaluate is: \{text\}
}}
\caption{Zero-shot scalar prompt for vision examples.}
\label{prompt:scalar_vision}
\end{figure*}

\begin{figure*}[t]
\noindent\fbox{\parbox{.98\textwidth}{\small\ttfamily
To help you make more accurate and consistent judgments, here is an expanded explanation of how to interpret and assign support percentages based on textual evidence. These examples are designed to cover a range of logic and linguistic relationships you may encounter.

100\% - Full and unambiguous support (Entailment): The sentence explicitly states the information in the claim, or the claim is a direct paraphrase of the sentence. There is no need for guessing; the facts are identical.

80-100\% - Almost complete support: The main assertions in the claim are present in the sentence. There may be minor differences in wording, synonyms, or omission of non-essential details, but the core meaning is fully preserved and supported.

60-80\% - Strong partial support (Strong Implication): The sentence strongly implies the claim is true through logical inference or context, though it may not state it explicitly. A reasonable person would conclude the claim is likely true based on the sentence.

40-60\% - Moderate partial support: There is a topical alignment or shared keywords. The sentence discusses the same subject matter, but the specific assertion in the claim is neither confirmed nor denied. It is plausible but lacks definitive evidence in the text.

20-40\% - Minimal weak support: There are weak textual links, such as matching entity names or a general theme, but the specific context is different. The sentence provides very little basis to deduce the claim.

0-20\% - Very weak or speculative support: There may be a very distant connection (e.g., related vocabulary), but inferring the claim from the sentence would be highly speculative.

0\% - No support or contradiction: The sentence is completely unrelated to the claim, or it directly contradicts the claim (proves it false).

Based on the provided sentence and claim, evaluate the probability that the claim is supported by the sentence. Your answer must be a decimal number between 0 and 1, and you must strictly follow the format below:

<answer>probability\_value</answer>

Where probability\_value is the result you calculate.

Sentence: \{sentence\}

Claim: \{claim\}
}}
\caption{Zero-shot scalar prompt for text examples.}
\label{prompt:scalar_text}
\end{figure*}

\begin{table*}[]
    \centering
    \setlength{\tabcolsep}{4pt}
\begin{tabular}{ll|ccc|ccc|cc}
\toprule
\multirow{4}{*}{Human} & \multirow{4}{*}{Judge} & \multicolumn{6}{c|}{\mirage} & \multicolumn{2}{c}{\multirow{3}{*}{Auto-ARGUE}} \\
\cmidrule(lr){3-8}
 & & \multicolumn{3}{c|}{\claimscore} & \multicolumn{3}{c|}{\citationscore} & \multicolumn{2}{c}{} \\
\cmidrule(lr){3-5}\cmidrule(lr){6-8}
 & & \multicolumn{2}{c}{\claimprec} & \multirow{2}{*}{\claimrec} & \multicolumn{2}{c}{\citeprec} & \multirow{2}{*}{\citerec} & \multicolumn{2}{c}{} \\
\cmidrule(lr){3-4}\cmidrule(lr){6-7}
 & & Ref. & Col. & & Ref. & Col. & & Sent. Supp. & Nugg. Cov. \\
\midrule
\multirow{2}{*}{Cite Supp.}
    & CLUE    & -$28.9$ & \phantom{-}$36.8$ & -- & \phantom{-}$8.1$ & \phantom{-}$71.3$ & -- & \multirow{2}{*}{\phantom{-}$66.3$} & \multirow{2}{*}{--} \\
    & Qwen3.5 & \phantom{-}$33.4$ & \phantom{-}$65.3$ & -- & \phantom{-}$53.3$ & \phantom{-}$\mathbf{79.2}$ & -- & & \\
\midrule
\multirow{2}{*}{Info Cov.}
    & CLUE    & -- & -- & \phantom{-}$73.7$ & -- & -- & \phantom{-}$64.5$ & \multirow{2}{*}{--} & \multirow{2}{*}{\phantom{-}$54.9$} \\
    & Qwen3.5 & -- & -- & \phantom{-}$\mathbf{77.7}$ & -- & -- & \phantom{-}$75.9$ & & \\
\bottomrule
\end{tabular}
    \caption{Kendall's Tau between \mirage, Auto-ARGUE, and RAGTIME human judgments.}
    \label{tab:ragtime_metric_combined}
\end{table*}

\begin{table*}[]
    \centering
\begin{tabular}{ccc|ccc|ccc}
\toprule
\multirow{3}{*}{Judge} & \multirow{3}{*}{Eval} & \multirow{3}{*}{Ano} & \multicolumn{3}{c|}{\claimscore} & \multicolumn{3}{c}{\citationscore} \\
\cmidrule(lr){4-6}\cmidrule(lr){7-9}
 & & & \multicolumn{2}{c}{\claimprec} & \multirow{2}{*}{\claimrec} & \multicolumn{2}{c}{\citeprec} & \multirow{2}{*}{\citerec} \\
\cmidrule(lr){4-5}\cmidrule(lr){7-8}
 & & & Ref. & Col. & & Ref. & Col. & \\
\midrule
\multirow{9}{*}{CLUE}
    & \multirow{4}{*}{EQJ}
        & 1 & \phantom{-}$19.7$ & \phantom{-}$44.2$ & \phantom{-}$64.0$ & \phantom{-}$3.3$ & \phantom{-}$6.3$ & \phantom{-}$26.3$ \\
    & & 2 & \phantom{-}$19.7$ & \phantom{-}$21.2$ & \phantom{-}$56.0$ & \phantom{-}$6.3$ & \phantom{-}$24.8$ & \phantom{-}$46.3$ \\
    & & 3 & \phantom{-}$34.8$ & \phantom{-}$36.3$ & \phantom{-}$82.3$ & \phantom{-}$13.3$ & \phantom{-}$8.2$ & \phantom{-}$32.7$ \\
    & & Avg & \phantom{-}$24.7$ & \phantom{-}$33.9$ & \phantom{-}$67.4$ & \phantom{-}$7.7$ & \phantom{-}$13.1$ & \phantom{-}$35.1$ \\
\cmidrule(l){2-9}
    & \multirow{4}{*}{ICJ}
        & 1 & \phantom{-}$30.0$ & \phantom{-}$23.3$ & \phantom{-}$58.2$ & \phantom{-}$16.7$ & \phantom{-}$11.5$ & \phantom{-}$41.8$ \\
    & & 2 & \phantom{-}$40.0$ & \phantom{-}$40.0$ & \phantom{-}$69.7$ & \phantom{-}$26.7$ & \phantom{-}$13.3$ & \phantom{-}$48.2$ \\
    & & 3 & \phantom{-}$40.0$ & \phantom{-}$26.7$ & \phantom{-}$69.7$ & \phantom{-}$26.7$ & \phantom{-}$13.3$ & \phantom{-}$53.3$ \\
    & & Avg & \phantom{-}$36.7$ & \phantom{-}$30.0$ & \phantom{-}$65.8$ & \phantom{-}$23.3$ & \phantom{-}$12.7$ & \phantom{-}$47.8$ \\
\cmidrule(l){2-9}
    & GJ & - & \phantom{-}$6.7$ & \phantom{-}$20.0$ & \phantom{-}$21.5$ & \phantom{-}$6.7$ & -$6.7$ & -$6.7$ \\
\midrule
\multirow{9}{*}{Qwen3.5}
    & \multirow{4}{*}{EQJ}
        & 1 & \phantom{-}$50.8$ & \phantom{-}$26.3$ & \phantom{-}$59.2$ & \phantom{-}$3.3$ & -$4.8$ & \phantom{-}$28.0$ \\
    & & 2 & \phantom{-}$38.2$ & \phantom{-}$14.5$ & \phantom{-}$84.2$ & \phantom{-}$31.5$ & -$19.7$ & \phantom{-}$49.3$ \\
    & & 3 & \phantom{-}$28.2$ & \phantom{-}$21.5$ & \phantom{-}$56.5$ & \phantom{-}$40.0$ & -$6.7$ & \phantom{-}$51.3$ \\
    & & Avg & \phantom{-}$39.1$ & \phantom{-}$20.8$ & \phantom{-}$66.6$ & \phantom{-}$24.9$ & -$10.4$ & \phantom{-}$42.9$ \\
\cmidrule(l){2-9}
    & \multirow{4}{*}{ICJ}
        & 1 & \phantom{-}$84.8$ & -$10.0$ & \phantom{-}$73.0$ & \phantom{-}$50.0$ & \phantom{-}$10.0$ & \phantom{-}$67.8$ \\
    & & 2 & \phantom{-}$66.7$ & \phantom{-}$0.0$ & \phantom{-}$72.7$ & \phantom{-}$40.0$ & -$6.7$ & \phantom{-}$63.0$ \\
    & & 3 & \phantom{-}$80.0$ & \phantom{-}$0.0$ & \phantom{-}$83.0$ & \phantom{-}$53.3$ & \phantom{-}$20.0$ & \phantom{-}$63.0$ \\
    & & Avg & \phantom{-}$77.2$ & -$3.3$ & \phantom{-}$76.2$ & \phantom{-}$47.8$ & \phantom{-}$7.8$ & \phantom{-}$64.6$ \\
\cmidrule(l){2-9}
    & GJ & - & -$6.7$ & \phantom{-}$73.3$ & -$13.3$ & -$20.0$ & \phantom{-}$26.7$ & -$6.7$ \\
\bottomrule
\end{tabular}
    \caption{Kendall's Tau between \mirage and \wikivideo human judgments. EQJ: Extrinsic Quality Judgment, ICJ: Intrinsic Claim Judgment, GJ: Grounding Judgment.}
    \label{tab:wikivideo_metric_combined}
    \vspace{-1em}
\end{table*}

\begin{table*}[]
    \centering
\begin{tabular}{cc|ccc|ccc}
\toprule
\multirow{3}{*}{Judge} & \multirow{3}{*}{Ano} & \multicolumn{3}{c|}{\claimscore} & \multicolumn{3}{c}{\citationscore} \\
\cmidrule(lr){3-5}\cmidrule(lr){6-8}
 & & \multicolumn{2}{c}{\claimprec} & \multirow{2}{*}{\claimrec} & \multicolumn{2}{c}{\citeprec} & \multirow{2}{*}{\citerec} \\
\cmidrule(lr){3-4}\cmidrule(lr){6-7}
 & & Ref. & Col. & & Ref. & Col. & \\
\midrule
\multirow{4}{*}{CLUE}
    & 1      & \phantom{-}$36.7$ & \phantom{-}$7.3$  & -$25.7$ & \phantom{-}$3.7$  & -$11.0$ & -$25.7$ \\
    & 2      & \phantom{-}$0.0$  & -$29.4$           & \phantom{-}$3.7$  & -$22.0$ & -$36.7$ & \phantom{-}$3.7$ \\
    & 3      & \phantom{-}$33.7$ & \phantom{-}$22.4$ & -$48.6$ & -$3.7$  & \phantom{-}$0.0$  & -$48.6$ \\
    & Avg    & \phantom{-}$23.4$ & \phantom{-}$0.1$  & -$23.5$ & -$7.4$  & -$15.9$ & -$23.5$ \\
\midrule
\multirow{4}{*}{Qwen3.5}
    & 1      & \phantom{-}$29.4$ & \phantom{-}$18.3$ & -$22.0$ & -$11.0$ & -$29.4$ & -$22.0$ \\
    & 2      & -$7.3$            & -$18.3$           & \phantom{-}$7.3$  & -$29.4$ & -$55.0$ & \phantom{-}$7.3$ \\
    & 3      & \phantom{-}$26.2$ & \phantom{-}$15.0$ & -$44.9$ & -$11.2$ & -$15.0$ & -$44.9$ \\
    & Avg    & \phantom{-}$16.1$ & \phantom{-}$5.0$  & -$19.9$ & -$17.2$ & -$33.1$ & -$19.9$ \\
\bottomrule
\end{tabular}
    \caption{Kendall's Tau (system-level, $\times 100$) between \mirage{} and MAGMaR human judgments, combining reference and collection evidence, for both judges. \claimprec{}/\citeprec{} are precision under reference (Ref.)\ and collection (Col.)\ evidence; \claimrec{}/\citerec{} are recall, scored against the reference, so a single value is reported. \claimscore/\citationscore label the claim and citation families and have no separate F1 column. MAGMaR's only human signal is a report-level preference score (RPJ). Avg is the mean of the per-annotator correlations; Pooled averages the annotator scores before ranking.}
    \label{tab:magmar_metric_combined}
    \vspace{-1em}
\end{table*}

\begin{table*}[]
    \centering
    \begin{tabular}{c|ccc|ccc|ccc}
    \toprule
         &  \multicolumn{3}{c}{ICJ 1} & \multicolumn{3}{c}{ICJ 2} & \multicolumn{3}{c}{ICJ 3} \\
         & F1 & Precision & Recall & F1 & Precision & Recall & F1 & Precision & Recall\\ 
         \midrule 
         EQJ 1 & $49.0$ & $37.8$ & $40.8$ & $65.3$ & $68.7$ & $59.0$ & $60.5$ & $36.0$ & $52.3$\\
         EQJ 2 & $61.3$ & $49.7$ & $68.0$ & $76.3$ & $53.0$ & $74.5$ & $81.2$ & $51.5$ & $81.2$ \\
         EQJ 3 & $54.5$ & $43.3$ & $64.2$ & $70.8$ & $57.8$ & $77.2$ & $72.7$ & $49.7$ & $74.2$ \\
    \bottomrule
    \end{tabular}
    \caption{Kendall's Tau agreement between \frameworkname human annotated judgments (ICJ[1-3]) and human annotated quality judgments (EQJ[1-3]).}
    \label{tab:human_metric_all}
\end{table*}

\begin{figure*}
\noindent\fbox{%
    \parbox{.98\textwidth}{%

\footnotesize
{\tt
\small
For each prediction, you should give a likert score from 1 to 5. The criteria for giving a score is as follows (in descending importance):

\begin{itemize}
    \item \textbf{Consistency/Factuality:} Does the article make only true statements about the topic in question, given what the reference says about that topic?
    \begin{itemize}
        \item Articles that make factual errors, omissions, or hallucinations of any kind should be penalized.
    \end{itemize}
    \item \textbf{Adequacy:} Does the article adequately capture all of the information contained in the reference article?
    \begin{itemize}
        \item Articles that omit details about any of the important aspects of the topic should be penalized.
    \end{itemize}
    \item \textbf{Coherence:} Does the article make sense on its own, as a standalone description of the topic?
    \begin{itemize}
        \item Articles that require you to go read the reference in order to understand what they mean (or that don't make sense even then) should be penalized.
        \item Articles that don't provide substance on the event should also be penalized (e.g. "the [event] happened at [a time] in [a place]")
    \end{itemize}
    \item \textbf{Relevancy:} Does the article include only information that is relevant to the topic in question?
    \begin{itemize}
        \item Articles that include irrelevant or superfluous information, or information about some topic other than the one represented by the reference article, should be penalized.
    \end{itemize}
    \item \textbf{Fluency:} Does the article sound reasonable natural (like something a native English speaker might actually write)?
    \begin{itemize}
        \item Articles that are disfluent or that sound unnatural should be penalized.
    \end{itemize}
\end{itemize}
Oftentimes, some of the summaries may be very similar to each other. It is totally fine to give multiple summaries the same score if you think they are of comparable quality!

You should enter your score for each summary in the score field. The default value for each summary is 0, meaning you have not yet annotated the likert judgment. Please do not use half scores (1.5, 2.5, etc).
}
}}

\caption{Instructions for the human likert judgments (EQJ)}
\label{prompt:human_likert_instructions}
\end{figure*}

\begin{figure*}
\noindent\fbox{%
    \parbox{.98\textwidth}{%

\footnotesize
{\tt
\small
In this task, you'll be given a set of 3 json files for each prediction and 3 json files for the reference. The reference json files will all be the same, but each will correspond to your annotations for a different system's predictions. In each json file, there will be the same 10 topics.

For every system prediction json, you'll see 
\begin{enumerate}
    \item \textbf{prediction:} the predicted article for the topic. This prediction is the model's attempt to write an article for the topic.
    \item \textbf{claims:} this is a sub dictionary that contains the claims and sentences they came from it is formatted below. The judgment is whether the claim is supported by the reference article or not.
    \begin{quote}
        \{
          "sentence": \{
            "claim": "the claim to be verified",
            "judgment": [False|True]
          \},
        \}
    \end{quote}
\end{enumerate}

\textbf{Annotation Task}

The goal of this task is to mirror how we evaluate multimodal RAG in InfoF1. The basic premise is as follows.

\begin{itemize}
    \item For every claim in the prediction, check if it is supported by the reference article. If it is mark the judgment as True, otherwise mark it as False. (The default value is None)
    \item For every claim in the reference, check if it is supported by the prediction article. If it is mark the judgment as True, otherwise mark it as False. (The default value is None)
\end{itemize}

\textbf{Claim Annotation Criteria}

When deciding whether a \textbf{claim is supported} by the other article (reference or prediction), apply the following criteria. The goal is to make judgments \textbf{only when there is no reasonable doubt} that the claim is supported by the other text.

A claim is \textbf{SUPPORTED (True)} if:
\begin{enumerate}
    \item \textbf{All factual elements} in the claim are explicitly stated or can be directly inferred from the other article without needing external knowledge.
    \begin{itemize}
        \item Example: Claim: ``The Eiffel Tower is located in Paris.'' $\rightarrow$ Supported if the other article states ``The Eiffel Tower is in Paris.''
    \end{itemize}
    \item The \textbf{meaning and intent} of the claim are \textbf{fully consistent} with the information in the article.
    \begin{itemize}
        \item Minor wording differences are acceptable if they don’t change meaning.
    \end{itemize}
    \item The \textbf{temporal or causal context} matches (e.g., dates, events, outcomes are consistent).
    \item If the claim includes \textbf{quantitative or categorical facts} (numbers, names, locations, affiliations), these details must exactly match what is stated in the other article.
\end{enumerate}

A claim is \textbf{NOT SUPPORTED (False)} if:
\begin{enumerate}
    \item The other article \textbf{contradicts} any part of the claim.
    \item The other article \textbf{omits or is ambiguous} about key details needed to verify the claim.
    \item The claim requires \textbf{inference beyond what’s stated}, such as outside knowledge, assumptions, or general reasoning not grounded in the text.
    \item The claim is \textbf{partially supported}, but not fully — i.e., some parts are correct while others are missing or uncertain.
\end{enumerate}

\textbf{General rule:}
\begin{quote}
    Only mark a claim as \textbf{True} (supported) if you can clearly point to a sentence or set of sentences in the other article that fully confirm it, leaving \textbf{no doubt} about its accuracy. Otherwise, mark it as \textbf{False} (not supported).
\end{quote}
}
}}

\caption{Instructions for the human metric judgments (ICJ)}
\label{prompt:human_metric_instructions}
\end{figure*}

\end{document}